\documentclass{vgtc}                          

\ifpdf
  \pdfoutput=1\relax                   
  \usepackage{graphicx}                
  \DeclareGraphicsExtensions{.pdf,.png,.jpg,.jpeg} 
\else
  \usepackage{graphicx}                
  \DeclareGraphicsExtensions{.eps}     
\fi%

\graphicspath{{figures/}{pictures/}{images/}{./}} 

\usepackage{microtype}                 
\PassOptionsToPackage{warn}{textcomp}  
\usepackage{textcomp}                  
\usepackage{mathptmx}                  
\usepackage{times}                     
\usepackage{cite}                      
\usepackage{tabu}                      
\usepackage{booktabs}                  
\usepackage[boxed,linesnumbered,commentsnumbered, ruled, vlined]{algorithm2e}

\onlineid{1365}

\vgtccategory{research}

\vgtcinsertpkg

\preprinttext{To appear in an IEEE VGTC sponsored conference.}

\usepackage[dvipsnames]{xcolor}
\usepackage{amsmath}

\title{Object Detection in the Context of Mobile Augmented Reality}

\author{Xiang Li\thanks{e-mail: xiang.li@oppo.com} %
\and Yuan Tian\thanks{e-mail: yuan.tian@oppo.com} %
\and Fuyao Zhang\thanks{email: fuyao.zhang@oppo.com} %
\and Shuxue Quan\thanks{email: shuxue.quan@oppo.com} %
\and Yi Xu\thanks{e-mail: yi.xu@oppo.com}}
\affiliation{OPPO US Research Center}

\abstract{In the past few years, numerous Deep Neural Network (DNN) models and frameworks have been developed to tackle the problem of real-time object detection from RGB images. Ordinary object detection approaches process information from the images only, and they are oblivious to the camera pose with regard to the environment and the scale of the environment. On the other hand, mobile Augmented Reality (AR) frameworks can continuously track a camera's pose within the scene and can estimate the correct scale of the environment by using Visual-Inertial Odometry (VIO). In this paper, we propose a novel approach that combines the geometric information from VIO with semantic information from object detectors to improve the performance of object detection on mobile devices. Our approach includes three components: (1) an image orientation correction method, (2) a scale-based filtering approach, and (3) an online semantic map. Each component takes advantage of the different characteristics of the VIO-based AR framework. We implemented the AR-enhanced features using ARCore and the SSD Mobilenet model~\cite{liu2016ssd} on Android phones. To validate our approach, we manually labeled objects in image sequences taken from 12 room-scale AR sessions. The results show that our approach can improve on the accuracy of generic object detectors by $12\%$ on our dataset.%
} 

\CCScatlist{
  \CCScatTwelve{Computer Graphics}{Graphics systems and interfaces}{Mixed / Augmented Reality}{};
\CCScatTwelve{Artificial intelligence}{Computer vision}{Computer vision problems}{Object detection};
}

\begin{document}
\teaser{
  \includegraphics[width=1.0\linewidth]{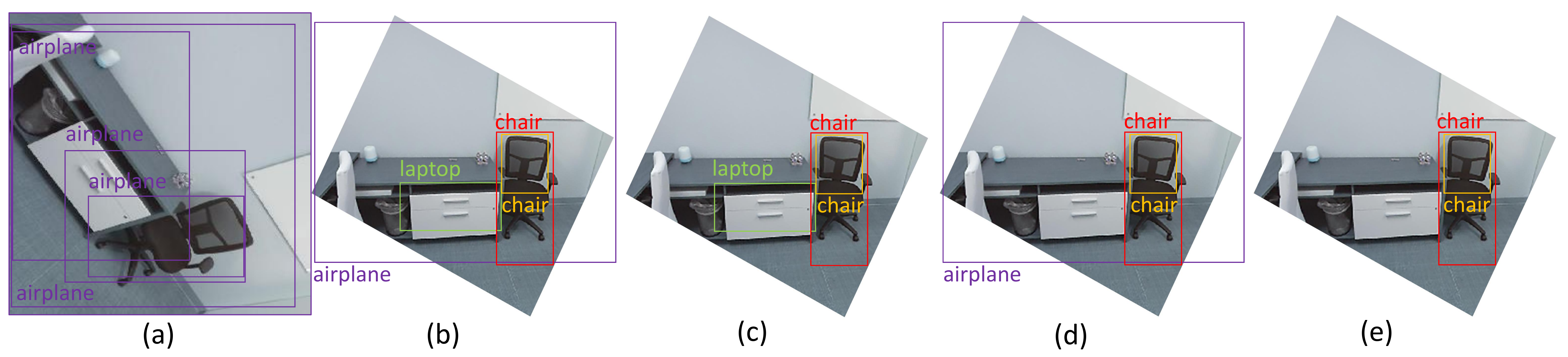}
   \caption{These images show the results of different object detection methods applied on the same image: (a) results of original DNN (SSD Mobilenet model), (b) DNN with image orientation correction, (c) DNN with image correction and scale-based filtering, (d) DNN with image correction and online semantic mapping, and (e) results of our proposed method.}
 \label{fig:result}
 
 }
\maketitle

\section{Introduction}

Object detection plays an important role in Augmented Reality (AR), where virtual content can be associated with real-world objects to enable applications such as product recommendations for e-commerce, and textual and graphical informational displays for educational purposes. In addition, the semantic information obtained by an object detector can be used to help VIO. Recently, tremendous progress has been made on object detection using Deep Neural Networks (DNNs) including SSD\cite{liu2016ssd}, YOLO\cite{redmon2016you}, faster R-CNN\cite{ren2015faster}, etc. With advancements in mobile computing power and lightweight DNN frameworks designed for mobile platforms 
(e.g., TensorFlow Lite), more and more object detection DNNs can run on smartphones in real-time\cite{wong2019yolo, howard2017mobilenets}. However, most of these general-purpose object detectors are not designed specifically for AR applications. First, most of them are trained with the objects oriented upright in the images, which presents challenges when the camera is rotated as seen in \autoref{fig:result}(a). This type of camera rotation frequently occurs when a user views virtual content during an AR session. For example, if a virtual object is large and wide, the user has a tendency to rotate the phone from a portrait to landscape orientation. Second, object detectors are oblivious to scale information due to the scale ambiguity of a single input image. Third, during an AR session, an object may be viewed from many different viewpoints. Since most object detectors do not memorize specific locations that the device has visited before, this can potentially result in inconsistent detection over a large range of viewpoints.

Although data augmentation and data synthesis can be used to ease the above-mentioned problems of scale ambiguity and viewpoint variance, it requires more training data and time. Another approach is to use the additional geometry information provided by a depth camera \cite{wang2019densefusion}. However, depth cameras on mobile devices are not widely available. Video-based methods can also alleviate the viewpoint variance problem (e.g.~\cite{han2016seq,liu2018mobile}),
but they are either limited to temporally-close frames or rely on intricate network designs. Our key insight is that we leverage the information generated by an AR framework, especially by a VIO method, to solve the aforementioned problems and improve the performance of object detectors in the context of mobile AR applications.

Mobile AR frameworks have become widely available for application developers, including Apple Inc.’s ARKit and Google Inc.’s ARCore. These AR frameworks use VIO to track a six degrees of freedom (6DoF) camera pose continuously during an AR session. These frameworks also provide sparse 3D points with real-world scales. In addition, AR frameworks have the ability to memorize places by tracking the camera within a map. Even when tracking is lost, re-localization can be used to re-estimate the camera pose with respect to the map. Intermittent re-localization is typically performed to detect revisited places in order to mitigate the drift accumulated in the camera trajectory over time.

\begin{figure*}[ht!]
 \centering 
 \includegraphics[scale=0.6]{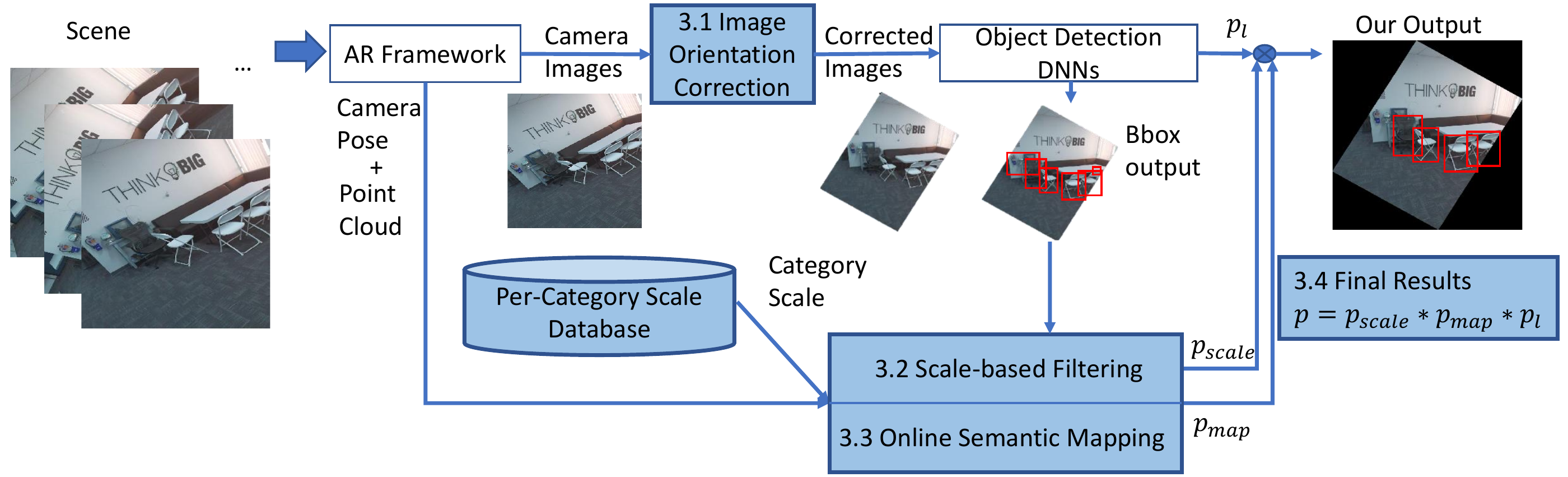}
 \caption{Overview pipeline of our approach. The different modules in our method are shown as blue boxes.}
 \label{fig:overview}
\end{figure*}

With recent advancements in mobile computing power, both AR frameworks and object detection DNNs can run simultaneously on a mobile device using a single camera as the input source. In this paper, a new object detection pipeline is proposed in the context of mobile AR. This pipeline takes advantage of the output and features of AR frameworks to increase the overall object detection accuracy of DNNs. Specifically, we use camera pose information from VIO to correct image orientation, use scale information to reject false positives, and build an online semantic map to fuse detection results from various viewpoints and distances on AR-capable mobile devices. We show that our approach can improve the accuracy of off-the-shelf object detection DNN models.


\section{Related Work}
In this section, we discuss related work on object detection, including object detection methods that handle scale and viewpoint variance, as well as SLAM methods that utilize semantic information. Finally, we show examples of how VIO output can benefit other computer vision algorithms.

\subsection{Object Detection}
Generic object detection
is one of the most fundamental computer vision problems. It provides a semantic understanding of the real world and is related to many applications. 
DNN-based object detection can be separated into two main types. The first type is region proposal followed by classification, such as 
Fast R-CNN~\cite{girshick2015fast}, Faster R-CNN~\cite{ren2015faster}, R-FCN~\cite{dai2016r}, FPN~\cite{lin2017feature} and Mask R-CNN~\cite{he2017mask}. The second type uses DNNs as direct regressors to produce the final categories and locations simultaneously. Examples of this second method include MultiBox~\cite{erhan2014scalable}, 
YOLO~\cite{redmon2016you}, SSD~\cite{liu2016ssd}, YOLOv2~\cite{redmon2017yolo9000},  DSSD~\cite{fu2017dssd}, etc.
R-CNN~\cite{girshick2014rich} was the first work that applied CNN-based feature extraction in the traditional object detection pipeline. Later, it was improved by adding spatial pyramid matching~\cite{he2015spatial}, by introducing multi-task loss on classification and bounding box regression~\cite{girshick2015fast}, by adding a Region Proposal Network (RPN)~\cite{ren2015faster}, and by adapting to a fully-convolutional architecture~\cite{dai2016r}. Mask R-CNN~\cite{he2017mask} adds an additional branch to predict the segmentation mask to enable instance segmentation. 
Many researchers have also investigated end-to-end frameworks for object detection. Redmon et al. proposed YOLO (you only look once)~\cite{redmon2016you} which uses the same feature map to predict both the bounding boxes and the confidence scores for categories. Another important work is SSD~\cite{liu2016ssd}, which uses anchor boxes with different aspect ratios and scales instead of fixed grids.
YOLO and SSD are two very popular frameworks and have continued to evolve in recent years~\cite{redmon2017yolo9000, fu2017dssd,wong2019yolo,lin2017focal}.  Recently, there has been research on 6D detection from RGB images\cite{xiang2015data, tekin2018real, tremblay2018deep}. This method can predict the pose of 3D objects from images but requires a lot of training data to do so. 
On mobile devices, Ahmadyan et al. ~\cite{ahmadyan2020instant} proposed a system that tracks the pose of an object in real-time for AR applications. Similar to early attempts at template-based object detection in AR \cite{uchiyama2012object} and map-based relocalization ~\cite{liu2019towards}, this 6D detection approach relies on a known category and/or geometry. Based on the tracking results, AR effects can be superimposed onto certain objects. Our work differs from previous methods in that we combine generic object detection with SLAM, so no prior knowledge of the scene or object is required.

\subsection{Scale Invariance and Consistency}
Although current object detection methods can achieve promising results on public datasets, there are still many challenges such as scale invariance, detection under large viewpoint changes, and detection consistency over time. 
To achieve scale invariance, image pyramids are used to extract multi-scale features~\cite{he2015spatial}. However, this requires a significant amount of additional training time and higher memory consumption. As a result, some DNNs only use pyramids in the testing phase~\cite{girshick2015fast, ren2015faster}. Feature hierarchy inside the DNNs has also been investigated for scale invariance ~\cite{liu2016ssd}. Multi-scale CNNs~\cite{cai2016unified} generate multiple output layers so that receptive fields match objects of different scales. 
Similarly, TridentNet \cite{li2019scale} generates scale-specific feature maps with a uniform representational power. HyperNet~\cite{kong2016hypernet} aggregates feature maps from different resolutions into a uniform space. FPN~\cite{lin2017feature} proposes a bottom-up and top-down architecture for the feature hierarchy. 
YOLO v2~\cite{redmon2017yolo9000} and SNIPER~\cite{singh2018sniper} also introduce multi-scale training. To improve object detection, researchers have also utilized multi-task learning~\cite{dai2016instance,brahmbhatt2017stuffnet} to boost object detection. Contextual information is also introduced to improve object detection performance using Markov Random Field~\cite{zhu2015segdeepm} and LSTMs~\cite{byeon2015scene}. Our work uses information from VIO to improve accuracy using off-the-shelf models without any additional training.

\subsection{Semantic SLAM}

With the development of deep learning, using semantic information for SLAM has drawn a great deal of attention. Many researchers ~\cite{mccormac2017semanticfusion,sunderhauf2017meaningful, yu2019variational} have proposed methods to combine semantic segmentation and SLAM together, where the semantic labels of objects are fused into the SLAM map to create a semantic map. MaskFusion ~\cite{runz2018maskfusion} proposed segmenting different objects in the scene while tracking and reconstructing them even as they move independently from the camera.  SLAM++~\cite{salas2013slam++} reconstructed the object-level environment with a 3D object database based on RGBD SLAM in real-time. Bowman et al. ~\cite{bowman2017probabilistic} built an object-level map and also utilized object association tightly coupled with SLAM to improve accuracy. Mu et al. ~\cite{mu2016slam} treated all object landmarks with discrete distribution to be coupled with SLAM. This was later improved by Zhang et al.~\cite{zhang2019hierarchical} by using a hierarchical Dirichlet process for data association. To the best of our knowledge, our work is the first to use VIO information to improve DNN-based object detection and provide a semantic map in real-time on AR capable mobile devices.

\subsection{Improving Algorithms using VIO}
AR capable mobile devices are often equipped with a camera and Inertial Measurement Unit (IMU) and use VIO to estimate camera pose. The additional sensor data and output of VIO can help solve challenging computer vision problems. Kurz et. al. \cite{kurz2011inertial} proposed gravity-aligned local feature descriptors for image classification. The idea is similar to our image orientation correction described in \autoref{section:rotationCor}.
For 3D reconstruction, Chisel \cite{klingensmith2015chisel} utilized RGBD data and camera pose information from VIO to reconstruct indoor scenes using an implicit Truncated Signed Distance Function data structure. For collision detection, Xu et. al.~\cite{xu2018multi} proposed a real-time multi-scale voxelization method using camera pose information and sparse point clouds from mobile AR frameworks. Visual occlusion can also be implemented by depth densification  ~\cite{holynski2018fast, tian2019occlusion} or by depth from motion \cite{valentin2018depth}.
Phalak et. al. ~\cite{indoorboundary} utilized a sequence of posed RGB images as input to a DNN to estimate the boundaries of the indoor environment. Our method takes advantage of VIO to improve object detection.

\section{Methodology}

\autoref{fig:overview} shows an overview of our pipeline. During a live AR session, a user moves around a scene to view virtual objects from different viewing angles while holding the phone using different orientations. The image sequence is consumed by the mobile AR framework (e.g., ARCore or ARKit), which uses VIO to track the pose of the mobile device in real-time. The pose is used to correct the image orientation so that the objects in the image appear to be in an upright orientation. The corrected images are passed to a DNN for object detection. 

Most DNNs used for object detection, such as YOLO, SSD, and Faster R-CNN, can output detection bounding boxes and corresponding labels. Although our method works with any of these pre-trained DNNs, we chose to focus on the efficient models that can be integrated with the AR framework on mobile devices. Moreover, since we are interested in indoor environments, we chose to use the SSDmobilenetV1coco model from the Tensorflow detection model zoo. The model is pre-trained on the COCO dataset\cite{lin2014microsoft}, which covers popular object categories in indoor environments. 

\begin{figure}[tb]
 \centering 
 \includegraphics[width=\columnwidth]{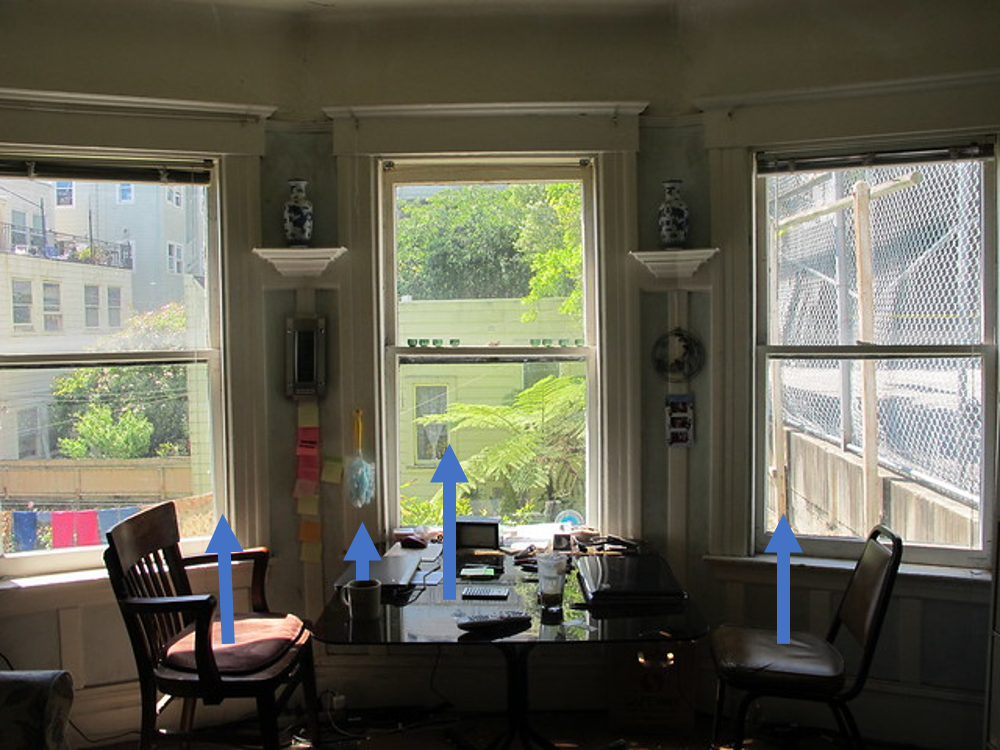}
 \caption{Most real-world objects are in an upright orientation, such as the desk, chair, and cup in this scene from the COCO dataset\cite{lin2014microsoft}.}
 \label{fig:upright}
\end{figure}

In addition to camera pose information, sparse point cloud data is typically provided by the AR framework as well. The point clouds and camera pose are combined with object detection output from the DNN into our scale-based filtering module and semantic map module. Our method improves object detection accuracy by taking advantage of the geometric information with real-world scale. The final detection result for the current frame is obtained by adjusting the DNN output using information from the two modules.



\subsection{Image Orientation Correction}
\label{section:rotationCor}

\begin{figure}[tb]
 \centering 
 \includegraphics[width=\columnwidth]{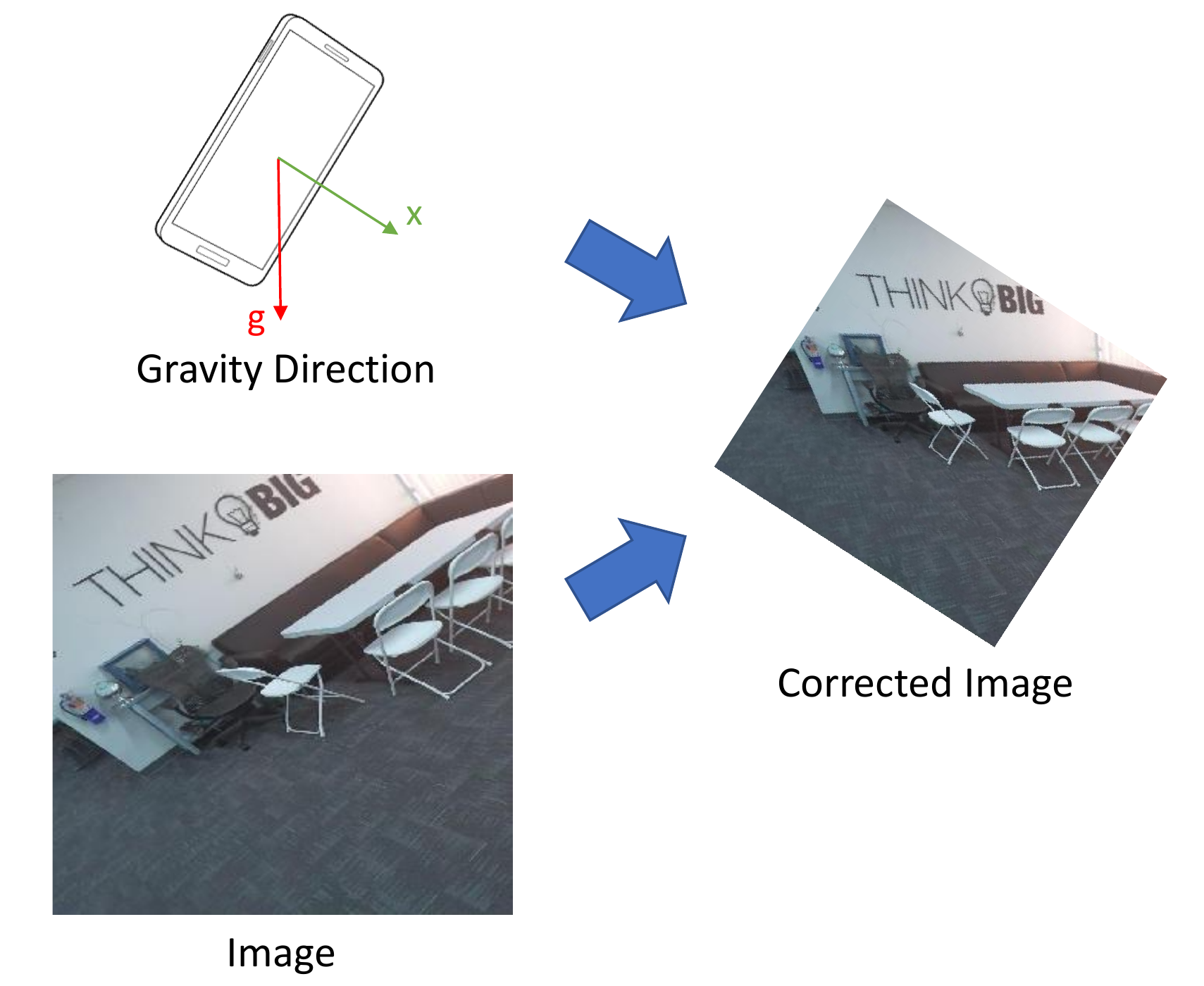}
 \caption{Image orientation correction using gravity direction.}
 \label{fig:rotation}
\end{figure}

If an out-of-the-box DNN-based object detection model is not trained with rotation augmentation, its detection capability will be impaired when the orientation of an object is not aligned with that of the camera. In this case, even if a detector has a certain level of built-in rotational invariance, the output bounding box of the object is still aligned with the camera axes and likely to include unwanted pixels from the background. 

We take advantage of the fact that most objects in real-world scenes are placed on horizontal surfaces and are in their upright orientations, as shown in \autoref{fig:upright}. One important feature of mobile devices is that the direction of gravity can be estimated by using the IMU sensor. For example, the Android platform provides the gravity sensor values that include the direction and magnitude of gravity. For an input image, it is straightforward to compute the angle to rotate the image around its center so that the \textit{x}-axis of the image is perpendicular to the direction of gravity. After this correction, objects that are placed on horizontal surfaces will appear in their upright orientations. In our pipeline, different from\cite{kurz2011inertial}, orientation correction works as a preprocessing step for the DNN model. 

This orientation correction can greatly increase the detection accuracy of certain object categories such as furniture and appliances. Furthermore, the object bounding boxes can be used to estimate the real-world vertical and horizontal size of an object in 3D space. This information is used in our scale-based filtering in the next step.

\subsection{Scale-based Filtering}
\label{section:scalematch}
The detection results from the object detector sometimes include false positives, many of which can be easily filtered if scale information is available (e.g., a book that is 5 m high is likely a false positive). Fortunately, most mobile AR frameworks provide a per-frame sparse 3D point cloud in a coordinate system with a real-world scale. In this section, we propose a method to estimate the scale of the objects within the scene and filter out false positives. 

\begin{figure}[tb]
 \centering 
 \includegraphics[scale=0.8, trim=0 40 0 40,clip]{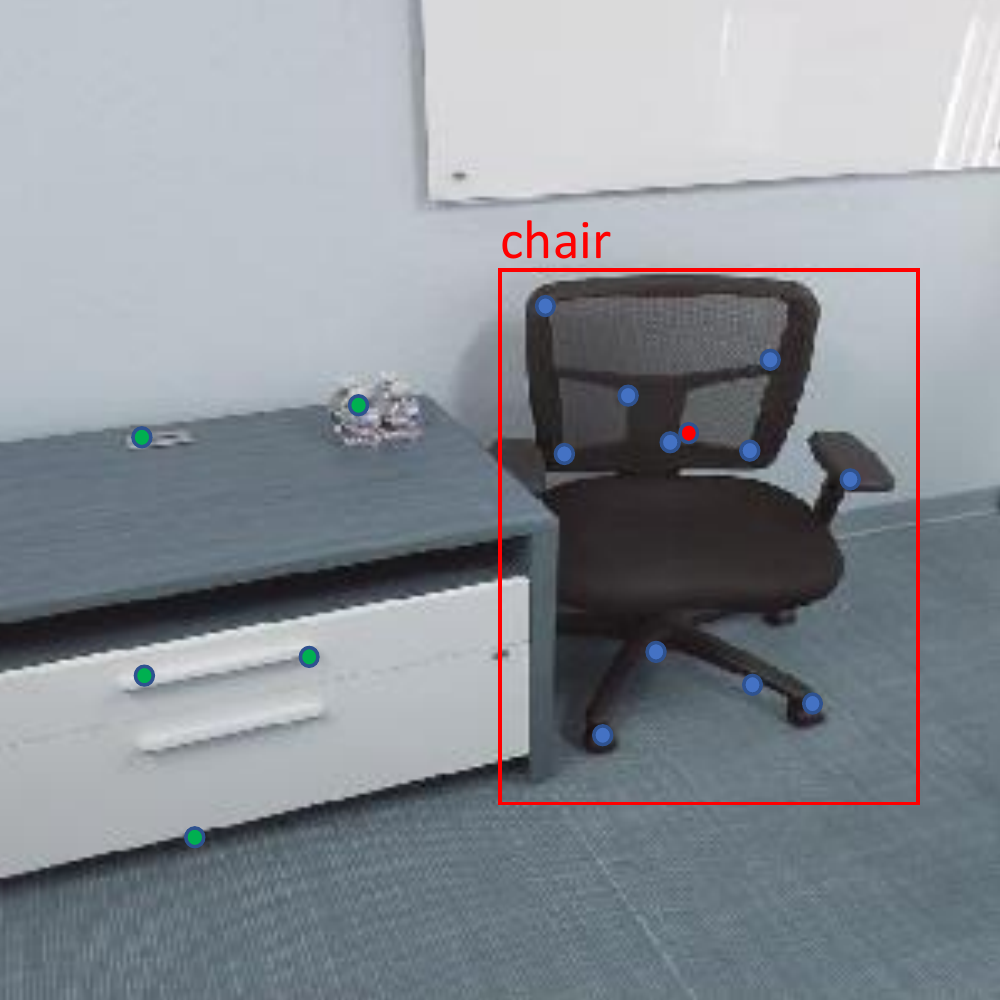}
 \caption{Scale estimation. The 3D sparse points are projected onto the image shown as dots. \textcolor{blue}{Blue dots} are points that are inside a 2D bounding box generated by the detector and the \textcolor{red}{red dot} is the projection of the median of all blue points.}
 \label{fig:scale}
\end{figure}

In each frame, the detector predicts bounding boxes around objects. For each box, we compute two values to represent an object's real scale in 3D using the following equation:

\begin{equation}
\label{eqn:3dsize}
\begin{aligned}
D_{w} = w/f_x * d \\
D_{h} = h/f_y * d
\end{aligned}
\end{equation}

where $D_{w}$ and $D_{h}$ are the estimated horizontal and vertical lengths of the object, $f_x$ and $f_y$ are focal lengths of the camera, $w$ and $h$ are the width and height of the 2D detection bounding box, and $d$ is the distance from the camera to the object. To compute $d$, we compute the median of the 3D sparse points produced by the AR framework whose 2D projections are within the detection bounding box (e.g., shown as blue dots in \autoref{fig:scale}). Then, $d$ is computed as the Euclidean distance between the camera position and the median. 

To filter out false positives based on the estimated scale, we introduce a per-category scale database including 80 categories defined in COCO. For each category, we manually define a minimum and maximum scale for the horizontal and vertical dimensions, resulting in four values for each category. \autoref{tab:scaleDatabase} provides a number of examples. Different object orientations may lead to varying widths of the 2D bounding box on an image (e.g., the width of a thin slab on an image will be very different from two different sides). Therefore, we use both the smallest and the largest possible values for the horizontal dimension when populating our database. Since the main goal of scale-based filtering is to reject false positives, using conservative values can ensure that true positives will be preserved during filtering.

The final step of scale-based filtering is to determine if the currently estimated scale matches the corresponding scale data from the database. Since both $D_{w}$ and $D_{h}$ are coarse estimations,  
we do not use the estimated dimension value to adjust the final detection probability. Instead, we only use these two values for filtering the false-positive results $p_{scale}$ as follows:

\begin{equation}
\label{eqn:scalefilter}
p_{scale} = 
\begin{cases}
1 &  if\ min_w \le D_w \le max_w \ and\  min_h \le D_h \le max_h \\
0.5 & otherwise
\end{cases}
\end{equation}

\begin{table}[tb]
  \caption{Example Data Entries in the Per-Category Scale Database}
  \label{tab:scaleDatabase}
  \centering
  \begin{tabular}{lllll}
  \hline
    
    Category            & $min_{w}$(m) & $max_{w}$(m) & $min_{h}$(m) & $max_{h}$(m) \\ 
    \hline
    chair     & 0.8   & 1.7          & 0.3 & 1.0 \\ \hline
    dining table    & 1.0   & 1.5    & 0.5 & 2.0 \\ \hline
    oven      & 0.3   & 1.4          & 0.4 & 1.2 \\ \hline
    refrigerator     & 0.5   & 2.0          & 0.5 & 1.4 \\ \hline
\end{tabular}
\end{table}

Scale-based filtering cannot distinguish between categories that have a similar scale range. A more general approach, semantic mapping, is proposed in the next subsection to take advantage of the multi-viewpoint data of an object typically available in an AR session.

\subsection{Online Semantic Mapping}
\label{section:semanticmap}

There are many challenges associated with object detection such as variance in scale and viewpoint. An object detector must detect objects with different scales on the images and from different viewpoints. Moreover, the object detector might produce inconsistent detection results over time. To overcome these problems, we build a semantic map using output from both the AR framework and the object detector. A probabilistic model is then used to maintain and update object category, viewpoint, and scale information within the semantic map. Finally, we extract information from the updated 3D semantic map to adjust the object label probability from the object detector for the current frame as shown in \autoref{fig:overview}. The idea is if an object can be correctly labeled by the detector most of the time over a range of different viewpoints, our approach should predict the correct label with higher confidence.



Our semantic mapping and updating process consists of three steps: (1) given the output of both the AR framework and the object detector, we create object point representations from the detection results on the current frame; (2) the new object points are fused with the superpoints that were already stored in the semantic map; and (3) the probability output from the object detector is updated. The entire algorithm is shown in \autoref{alg:semantic}.

\begin{algorithm}
	\small
	\SetNlSty{small}{}{:}
	\While{object detection is active}
    {
		\For {each object detected on the current frame}
        {
            create one object point with the representation $(loc^{in}, l^{in}, v^{in}, s^{in})$;\\
            find all superpoints whose distance to the incoming object is within the maximum scale of object category $l^{in}$ and insert them into a set $S^{in}$. \\
            \If{$S^{in}$ is not empty}{
        			\For {each superpoint $sp_i$ in $S^{in}$}
        			{
        			    update the score $E_l^{sp}$ using \autoref{eqn:elsp}\\
        			    $list\_{view}\leftarrow v^{in}$ if $v_{diff}\geq45^{\circ}$\\
        			    $list\_scale \leftarrow s^{in}$ if $s_{diff}\geq1$\\
        			}
			}
		    \If{no superpoints within the minimum scale distance of the incoming object point}{
			    Add a new superpoint with the representation\\
                $(loc\leftarrow loc^{in} , 
                list\_score(E^{in}_l), 
                list\_view(v^{in}), list\_scale(s^{in}))$\\
            }
			update the probability of current object using \autoref{eqn:final} 
		}
	}
\caption{Online Semantic Mapping.}\label{alg:semantic}
\end{algorithm}

\textbf{Step 1:} For each frame, the object detector outputs a list of $N$ object categories with associated bounding boxes and probabilities. For each object, we create one Object Point with the representation $(loc, label, view, scale)$, where $loc$ is the 3D coordinates of the median of all sparse points within the bounding box (same as in the scale estimation step of \autoref{section:scalematch}), $label$ is the object label, $view$ is the view direction from the camera to $loc$ and is a normalized unit vector, and $scale$ is the scale information that depends on the distance $d$ from camera position to $loc$. The scale information $s$ is an integer defined as:

\begin{equation}
\label{eqn:scale}
s = \lfloor{log}_2d\rceil
\end{equation}





\textbf{Step 2:} The second step is to fuse the estimated object with the information already in the semantic map, which stores a set of superpoints for objects in the scene. Different from the estimated object point data structure, the object superpoints are represented as $(loc, list\_score, list\_view, list\_scale)$. The three lists encode information from all previous frames in the same AR session. 
\begin{itemize}
    \item  $list\_score(E_1, E_2, E_3, \cdots, E_l,\cdots)$ is a list of the scores $E_l$ for each label $l$ that has been detected at this superpoint location. The higher the score, the higher probability this superpoint is of category $l$. The initial score values for all labels are 0. 
    \item $list\_view(v_1, v_2, v_3, \cdots)$ is a list of historical view directions from camera positions to the superpoint when an object of any category is detected at the superpoint.
    \item $list\_scale(s_1, s_2, s_3, \cdots)$ is a list of historical scales when an object of any category is detected at the superpoint.  
\end{itemize}

It is worth noting that different median locations ($loc$) from different viewpoints during an AR session might be computed for the same object because each image only shows a partial surface of an object. Moreover, the same object might be given different labels at different time instances. To fuse each incoming estimated object point $({loc}^{in},\ {l}^{in},\ {v}^{in},\ {s}^{in})$ with the superpoints in the semantic map, we first find a set of superpoints $S^{in}$ whose $loc$ are within a threshold distance to the incoming object point ${loc}^{in}$. We use the maximum scale of category $l^{in}$ from the scale database as the threshold when selecting set $S^{in}$. 

For the incoming object point of category $l$, we compute its score $E_l^{in}$ as follows:

\begin{equation}
\label{eqn:ein}
E_l^{in}={\frac{w_v+w_s}{2}\ast\ p}_{l}
\end{equation}

where $p_l$ is the probability of the category $l$ generated by the DNN and weights $w_v$ and $w_s$ are calculated as: 

\begin{equation}
\label{eqn:wv}
w_v = 
\begin{cases}
0                         & {if\ \ v_{diff} < 45^\circ }\\
(v_{diff}-45)\ /\ 45     & {if\ \ 45^\circ\le v_{diff}\le 90^\circ}\\
1                        & {otherwise}
\end{cases}
\end{equation}

\begin{equation}
\label{eqn:ws}
w_s = 
\begin{cases}
k_s\ast s_{diff}            & {if\ \ s_{diff} < 1/k_s}\\
1                        & {otherwise}
\end{cases}
\end{equation}

where $v_{diff}$ is the minimum absolute angular difference between $v^{in}$ and all view directions in the $list\_view$s of all points in $S^{in}$. The higher the $v_{diff}$, the higher the weight $w_v$ is. In this way, we value information observed from different viewing angles. We also set the weight $w_v$ to zero when the $v_{diff}$ is smaller than 45 degrees to only update the semantic map intermittently. $w_v$ is capped at 1 when the $v_{diff}$ is larger than 90 degrees. 

Similarly, $s_{diff}$ is the minimum absolute scale difference between $s^{in}$ and all scales in the $list\_scale$s of all points in $S^{in}$. The higher the $s_{diff}$, the higher the weight $w_s$ is. $k_s$ is used to normalize $s_{diff}$ based on a value range. 
In our method, this range is set empirically to be $[0, 5]$. Therefore $k_s$ is set to be $0.2$.

If the set $S^{in}$ is not empty, for each superpoint $sp\in S^{in}$, we update its detection score $E_l^{sp}$ using the information of the new detection of category $l$. We first check if the score $E_l^{sp}$ exists in $list\_score$. If not, we initialize $E_l^{sp}$ with $0$ and add it into $list\_score$. Then we update $E_l^{sp}$ as follows:

\begin{equation}
\label{eqn:elsp}
E_l^{sp} = 
\begin{cases}
E_l^{sp}+E_l^{in}            & {if\ \ maxScale\geq D>minScale }\\
E_l^{sp}+E_l^{in}+1          & {if\ \ D\le minScale}
\end{cases}
\end{equation}
where $D$is the distance  between the incoming point and $sp$. If $sp$ is within the minimum scale of the incoming object, we add 1 to the score of the detected label as a reward.

We also add view direction of the current detection $v^{in}$ into $list\_view$ of $sp$ if $v_{diff}\geq45^{\circ}$. We add scale of current detection $s^{in}$ into $list\_scale$ of $sp$ if $s_{diff}\geq1$. 

If there is no superpoint within the minimum distance of the incoming object point, we initialize the three lists $list\_score$, $list\_view$ and $list\_scale$ with the values of $E_l^{in}$, $v^{in}$ and $s^{in}$. Then we add a new superpoint with these three lists into the map.
\\

\textbf{Step 3}:  For each incoming object point $p$ with label $l$, we update its label probability $p_l$ using the nearby superpoints stored in the semantic map. We find a second set of superpoints whose $loc$ are within a distance (minimum scale of category $l^{in}$) to the incoming object point ${loc}^{in}$. We call this $S^{update}$. We find the maximum score $E_l^{max}$ among all the superpoints with label $l$ in $S^{update}$. We then find the maximum score $E_{\bar{l}}^{max}$ among all the superpoints with any other label in $S^{update}$. The probability $p_{map}$ of the semantic map at point $p$ is defined by a modified sigmoid function as follows:

\begin{equation}
\label{eqn:pmap}
p_{map} = 
\begin{cases}
\frac{1}{1+e^{E_{\bar{l}}^{max}-E_l^{max}}}      & {if E_l\geq E_{\bar{l}} }\\
0.5                                            & {if E_l<E_{\bar{l}}}
\end{cases}
\end{equation}

$p_{map}$ is at least 0.5 to guarantee that it does not decrease the output probability $p_l$ from the detector dramatically. 

\subsection{Final Detection Results}
Combining scale-based filtering and adjustment using semantic map, the final probability of an object is:

\begin{equation}
\label{eqn:final}
p=p_{scale}\ast p_{map}\ast p_l
\end{equation}

For each frame, the final output is a list of bounding boxes, each of which has the output $(l,\ p,\ bbox)$, where the label and bounding box are the same as the original output from the DNN. 

\section{Evaluation}
In this section, we show the evaluation of the proposed method compared with detection results from the DNN only.

\subsection{Datasets}

Our approach does not need extra data for re-training the DNN. Unfortunately, none of the popular image datasets for object detection provide the 3D sparse point cloud and camera pose information acquired by an AR framework. For this paper, we collected a new indoor dataset for evaluation. 

\begin{figure}[tb]
 \centering 
 \includegraphics[width=\columnwidth]{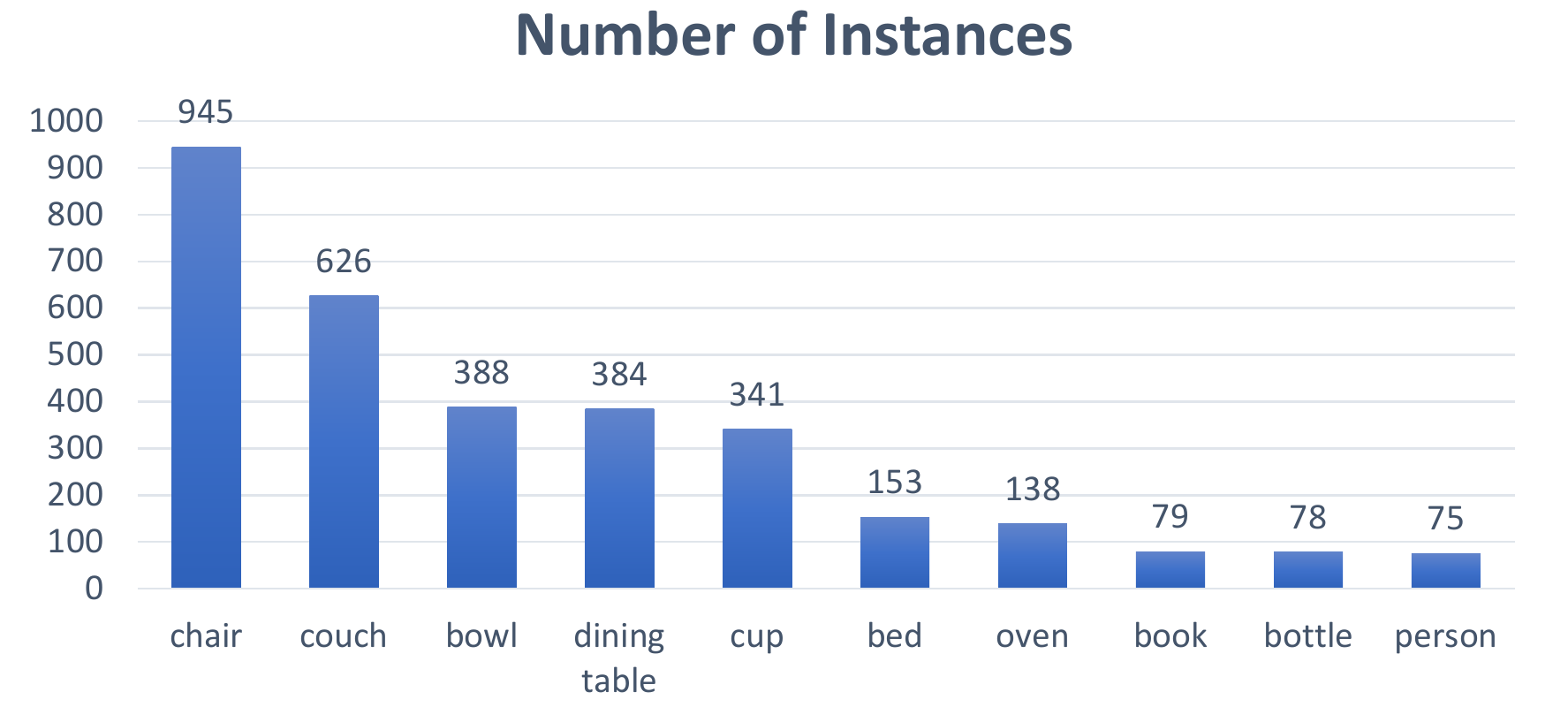}
 \caption{Number of instances for each of the top 10 categories of our dataset.}
 \label{fig:insStat}
\end{figure}

\begin{figure}[tb]
 \centering 
 \includegraphics[width=\columnwidth]{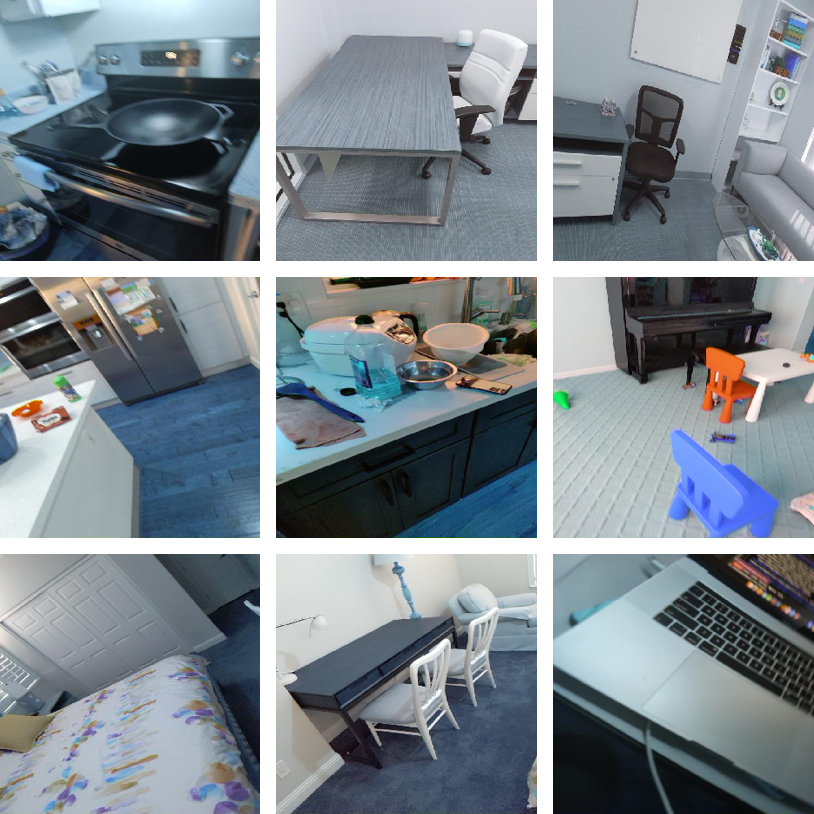}
 \caption{Examples of our testing dataset}
 \label{fig:dataset}
\end{figure}

Our indoor dataset includes about 2,000 images with 20 object categories and 3,384 instances in total, primarily obtained from scanning home and office environments. It also includes per-frame camera pose and sparse point cloud information from 12 room-scale AR sessions. \autoref{fig:insStat} shows the number of instances in the top 10 categories of our dataset. The remaining 10 categories in our dataset have a total of $167$ instances. Chairs were the most commonly scanned object, and the chair category is the largest consisting of 945 instances. Some example images are shown in \autoref{fig:dataset}. We use a mobile phone app developed using ARCore for data acquisition. The app collects images at about 1 fps while ARCore is running continuously to track the camera pose of the device. Each session is about 1 to 2 minutes long. We encouraged the data collectors to capture objects from various viewpoints to simulate real-world AR use cases where users tend to observe virtual content from different viewing angles. The collected images are also rotation corrected by using the method introduced in \autoref{section:rotationCor} and used to compare our method with SSD (MobileNet) in an ablation study where rotation correction is removed. We labeled objects in the original and corrected images using the categories of COCO dataset\cite{lin2014microsoft}. 

\subsection{Results}
\label{section:results}

We tested all modules of our approach on our dataset using the SSD\_mobilenet\_V1\_coco model from the Tensorflow detection model zoo. \autoref{fig:result} shows the visualization of results on one uncorrected example image using different methods. The original DNN (SSD MobileNet) result in \autoref{fig:result}(a) has false positives because of the insufficient data augmentation for training. \autoref{fig:result}(b) shows that our image orientation correction process can ease this problem.  \autoref{fig:result}(c) shows the result after applying scale-based filtering: the "airplane" label is filtered out as a false positive. \autoref{fig:result}(d) shows the result of online semantic mapping: the "laptop" label is also removed by the information from the semantic map. \autoref{fig:result}(e) shows the result using our proposed method, which combines all of the modules and generates a more accurate result.

\begin{figure}[tb]
 \centering 
 \includegraphics[width=\columnwidth]{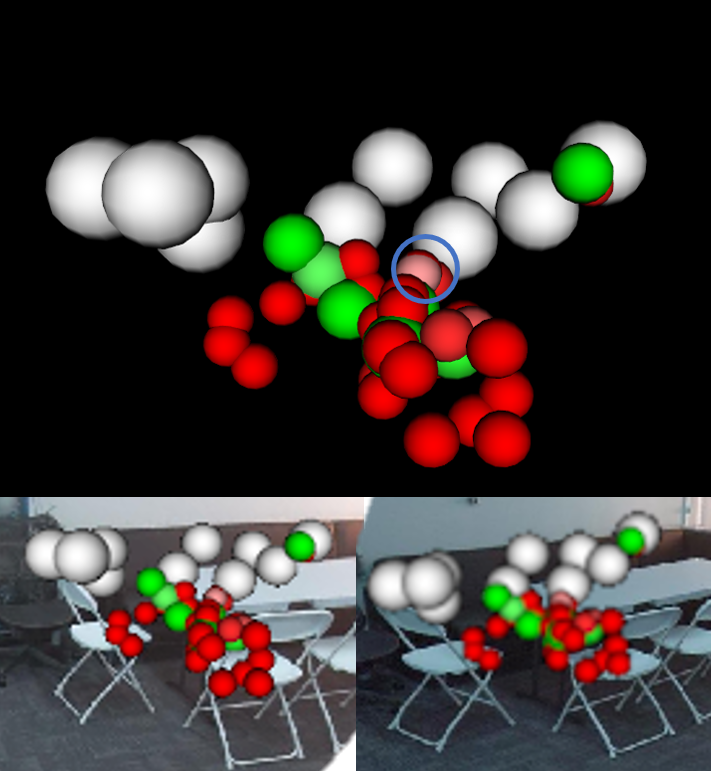} \caption{Visualization of a semantic map. Each sphere represents a superpoint. White is for the couch (note that there is a black couch behind the table), red is for the chairs, and green is for the dining table. The bottom two images visualize superpoints on two different frames. }
 \label{fig:semantic}
\end{figure}

To validate our online semantic mapping module, a semantic map is visualized in \autoref{fig:semantic}. The semantic superpoints are shown as spheres on two images from different viewpoints. The radius of a sphere reflects the minimum scale of the corresponding object category. Different colors represent different categories: white for the couch, red for the chair, and green for the dining table. As shown in \autoref{fig:semantic}, the superpoints maintain the correct semantic information. The visualization only shows the class label with the highest score for each superpoint, and the color saturation represents the score value. A blue circle highlights one superpoint which has a relatively low score for the current class label. It shows that using our semantic mapping, a superpoint that is in proximity to other superpoints with different labels tends to have a lower score. The semantic map is updated online, and it provides a more accurate probability for object detection. 

\autoref{tab:apresults} shows the evaluation results using metrics used by COCO. In \autoref{tab:apresults}, SSD represents the results of the original SSD MobileNet model, OC represents results using image orientation correction, SF represents results using scale-based filtering, and OSM represents results using online semantic mapping. As the results show, each of our modules has generated higher average precision compared with the original SSD MobileNet model. In particular, our entire pipeline ALL (OSM+SF+OC+SSD) has increased the average precision from $8.0\%$ to $20.4\%$. \autoref{tab:apresults} also shows that scale-based filtering and online semantic mapping improve detection results in different ways, since the average precision of our overall method ($20.4\%$) is significantly higher than those of SF+OC+SSD ($14.6\%$)
and OSM+OC+SSD ($11.3\%$). 

\autoref{tab:arresults} shows that all individual modules improve the average recall, as well, although the average recall of the entire pipeline has decreased from the original model by $4\%$. This is because only the bounding boxes that can survive both modules will be accepted by our pipeline. This results in the missed detection of new objects or truncated objects. The definitions of superscripts of average precision and average recall metrics in both \autoref{tab:apresults} and \autoref{tab:arresults} are the same as those in the COCO dataset\cite{lin2014microsoft}.

We have also tested Google MediaPipe 6D object detection~\cite{ahmadyan2020instant} on our dataset. To compare our 2D detection bounding boxes with MediaPipe's 3D bounding boxes, we compute a 2D bounding box based on each output cuboid of 6D detection. We set the IOU threshold as 0.5 for evaluation purposes. Using a larger threshold could be seen as unfair to 6D detection since the 2D box always covers a larger area than the corresponding cuboid projected on the image. We tested their chair model since the chair model is one of the available models and accounts for almost one third of the instances in our dataset. Our results show that MediaPipe has a $14.4\%$ average precision on our dataset. The results also indicate that our dataset, which is captured from real homes and offices, is very challenging.

All of the results were tested on a Pixel 3 smartphone. Our proposed method can run at $\sim$ 7 fps while ARCore tracks the pose of the camera in real-time. In comparison, the SSD inference alone runs at $\sim$ 7 fps while ARCore is running, confirming that our method is very efficient. We also expect that our method can build a semantic map for large scale environments since we only store a small number of superpoints for each detected object instead of a point cloud or voxels. The selected SSD model takes $300\times300$ image as input and outputs the top 10 bounding boxes with labels and properties. 

\begin{table}[tb]
  \caption{AP Metrics}
  \label{tab:apresults}
  \centering
  \begin{tabular}{lllllll}
  \hline
    Data & $AP$ & $AP^{.5}$ & $AP^{.75}$ & $AP^{s}$ & $AP^{m}$  & $AP^{l}$      \\ \hline
    SSD         &8.0  & 10.3  & 7.4  & 0.28       & 5.6         & 14.7        \\ \hline
    OC+SSD          &10.7 & 13.8  & 10.0 & 0.75       & 11.0        & 17.6        \\ \hline
    SF+OC+SSD       &14.6 & 18.9  & 13.5 & 0.60       & 14.6        & 19.1        \\ \hline
    OSM+OC+SSD &11.3 & 14.7  & 10.6 & 0.63       & 12.8        & 20.3        \\ \hline
    ALL         &20.4 & 26.3  & 18.9 & 0.78       & 19.0        & 26.0        \\ \hline
\end{tabular}
\end{table}

\begin{table}[tb]
  \caption{AR Metrics}
  \label{tab:arresults}
  \centering
  \begin{tabular}{lllll}
  \hline
    Data  & $AR^{10}$ & $AR^{s}$ & $AR^{m}$ & $AR^{l}$ \\ \hline
    SSD         &24.0         & 0.1        & 6.3         & 17.6       \\ \hline
    OC+SSD          &30.0         & 0.5        & 11.1        & 18.3       \\ \hline
    SF+OC+SSD       &24.6         & 0.0        & 9.6         & 15.0       \\ \hline
    OSM+OC+SSD         &26.0         & 0.4        & 9.6         & 16.0       \\ \hline
    ALL         &20.0         & 0.0        & 7.4         & 12.5       \\ \hline
\end{tabular}
\end{table}

\section{Conclusion}

There is a large amount of ongoing research on 2D object detection DNNs, and some of them have achieved high efficiency and great performance on mobile devices. Compared with 6D object detection \cite{wang2019densefusion}, it is much easier to collect and label 2D object boxes. Instead of 6D object detection, we take a hybrid approach where we utilize the 3D perception capabilities of the AR framework to improve 2D object detection. In this paper, we proposed three modules to achieve this purpose: we introduced an image orientation correction method to improve the input data to the object detector, we also introduced a per-category scale database as \textit{a prior} knowledge to filter 2D detection results based on their estimated real-world scales, and finally we proposed an online semantic mapping approach to further improve detection accuracy. Our approach works with any object detector as long as the output includes object bounding boxes, labels, and probabilities. However, it is challenging to integrate other DNN models with ARCore or ARkit while maintaining real-time performance, with the exception of the SSD models maintained by Google. We plan to keep working on such integration in the future. 



\section{Limitations and Future Work}
One limitation of our approach is that a newly detected truncated object might get rejected due to bad scale estimation and the fact that our map has no prior information about the detection. This can happen when the object is occluded by other objects or is only partially imaged by the camera. However, our comprehensive approach may still work if the online semantic map is continuously being updated using the output from the object detector. Another limitation is that the 3D sparse point clouds generated by the AR framework do not have correspondence between frames. Moreover, the 3D points do not have temporal consistency due to optimization during the SLAM process, which reduces the quality of our semantic map. In the future, we would like implement our own VIO algorithm so that we can further leverage the output of the VIO. Finally, our work may fail when the object being scanned moves.

There are a few future work directions that we can pursue. In this paper, we only use the sparse point cloud provided by the AR framework and bounding boxes from the object detector. In the future, we would like to explore how our pipeline can benefit from instance segmentation and dense depth map input. We would also like to evaluate how the accuracy of the VIO system under various conditions (e.g., scene depth, illumination, etc.) affects our object detection method. Moreover, we would like to extend our per-category scale database to make our approach more generic. For example, we would like to explore how to handle dynamic objects better by adding movement information to the database. Another idea is to customize the database specifically for the known environment in order to make the scale-based filtering more accurate.

\acknowledgments{
The authors wish to thank the reviewers for their detailed feedback and suggestions. We would also like to thank our colleagues at the OPPO US Research Center that helped us capture images of their homes. Finally, we wold like to thank Chris Vick for helping with paper writing.}

\bibliographystyle{abbrv-doi}

\bibliography{template,objectdetection,ar,slam}

\begin{thebibliography}{10}

\bibitem{ahmadyan2020instant}
A.~Ahmadyan, T.~Hou, J.~Wei, L.~Zhang, A.~Ablavatski, and M.~Grundmann.
\newblock Instant 3{D} object tracking with applications in augmented reality.
\newblock In {\em IEEE/CVF Conference on Computer Vision and Pattern
  Recognition (CVPR), Fourth Workshop on Computer Vision for AR/VR}, 2020.

\bibitem{bowman2017probabilistic}
S.~L. Bowman, N.~Atanasov, K.~Daniilidis, and G.~J. Pappas.
\newblock Probabilistic data association for semantic {SLAM}.
\newblock In {\em IEEE International Conference on Robotics and Automation
  (ICRA)}, pp. 1722--1729, 2017.

\bibitem{brahmbhatt2017stuffnet}
S.~Brahmbhatt, H.~I. Christensen, and J.~Hays.
\newblock Stuff{N}et: Using ‘stuff’to improve object detection.
\newblock In {\em IEEE Winter Conference on Applications of Computer Vision
  (WACV)}, pp. 934--943, 2017.

\bibitem{byeon2015scene}
W.~Byeon, T.~M. Breuel, F.~Raue, and M.~Liwicki.
\newblock Scene labeling with {LSTM} recurrent neural networks.
\newblock In {\em IEEE Conference on Computer Vision and Pattern Recognition
  (CVPR)}, pp. 3547--3555, 2015.

\bibitem{cai2016unified}
Z.~Cai, Q.~Fan, R.~S. Feris, and N.~Vasconcelos.
\newblock A unified multi-scale deep convolutional neural network for fast
  object detection.
\newblock In {\em European conference on computer vision (ECCV)}, pp. 354--370,
  2016.

\bibitem{dai2016instance}
J.~Dai, K.~He, and J.~Sun.
\newblock Instance-aware semantic segmentation via multi-task network cascades.
\newblock In {\em IEEE Conference on Computer Vision and Pattern Recognition
  (CVPR)}, pp. 3150--3158, 2016.

\bibitem{dai2016r}
J.~Dai, Y.~Li, K.~He, and J.~Sun.
\newblock {R-FCN}: object detection via region-based fully convolutional
  networks.
\newblock In {\em 30th International Conference on Neural Information
  Processing Systems}, pp. 379--387, 2016.

\bibitem{erhan2014scalable}
D.~Erhan, C.~Szegedy, A.~Toshev, and D.~Anguelov.
\newblock Scalable object detection using deep neural networks.
\newblock In {\em IEEE Conference on Computer Vision and Pattern Recognition
  (CVPR)}, pp. 2147--2154, 2014.

\bibitem{fu2017dssd}
C.-Y. Fu, W.~Liu, A.~Ranga, A.~Tyagi, and A.~C. Berg.
\newblock {DSSD}: deconvolutional single shot detector.
\newblock {\em arXiv preprint arXiv:1701.06659}, 2017.

\bibitem{girshick2015fast}
R.~Girshick.
\newblock Fast {R-CNN}.
\newblock In {\em IEEE International Conference on Computer Vision (ICCV)}, pp.
  1440--1448, 2015.

\bibitem{girshick2014rich}
R.~Girshick, J.~Donahue, T.~Darrell, and J.~Malik.
\newblock Rich feature hierarchies for accurate object detection and semantic
  segmentation.
\newblock In {\em IEEE Conference on Computer Vision and Pattern Recognition
  (CVPR)}, pp. 580--587, 2014.

\bibitem{han2016seq}
W.~Han, P.~Khorrami, T.~L. Paine, P.~Ramachandran, M.~Babaeizadeh, H.~Shi,
  J.~Li, S.~Yan, and T.~S. Huang.
\newblock Seq-{NMS} for video object detection.
\newblock {\em arXiv preprint arXiv:1602.08465}, 2016.

\bibitem{he2017mask}
K.~He, G.~Gkioxari, P.~Doll{\'a}r, and R.~Girshick.
\newblock Mask {R-CNN}.
\newblock In {\em IEEE International Conference on Computer Vision (ICCV)}, pp.
  2980--2988, 2017.

\bibitem{he2015spatial}
K.~He, X.~Zhang, S.~Ren, and J.~Sun.
\newblock Spatial pyramid pooling in deep convolutional networks for visual
  recognition.
\newblock {\em IEEE Transactions on Pattern Analysis and Machine Intelligence},
  37(9):1904--1916, 2015.

\bibitem{holynski2018fast}
A.~Holynski and J.~Kopf.
\newblock Fast depth densification for occlusion-aware augmented reality.
\newblock {\em ACM Transactions on Graphics (TOG)}, 37(6):1--11, 2018.

\bibitem{howard2017mobilenets}
A.~G. Howard, M.~Zhu, B.~Chen, D.~Kalenichenko, W.~Wang, T.~Weyand,
  M.~Andreetto, and H.~Adam.
\newblock Mobilenets: Efficient convolutional neural networks for mobile vision
  applications.
\newblock {\em arXiv preprint arXiv:1704.04861}, 2017.

\bibitem{klingensmith2015chisel}
M.~Klingensmith, I.~Dryanovski, S.~Srinivasa, and J.~Xiao.
\newblock Chisel: Real time large scale 3d reconstruction onboard a mobile
  device using spatially hashed signed distance fields.
\newblock In {\em Robotics: Science and Systems}, vol.~4, p.~1, 2015.

\bibitem{kong2016hypernet}
T.~Kong, A.~Yao, Y.~Chen, and F.~Sun.
\newblock Hyper{N}et: Towards accurate region proposal generation and joint
  object detection.
\newblock In {\em IEEE Conference on Computer Vision and Pattern Recognition
  (CVPR)}, pp. 845--853, 2016.

\bibitem{kurz2011inertial}
D.~Kurz and S.~B. Himane.
\newblock Inertial sensor-aligned visual feature descriptors.
\newblock In {\em IEEE Conference on Computer Vision and Pattern Recognition
  (CVPR)}, pp. 161--166, 2011.

\bibitem{li2019scale}
Y.~Li, Y.~Chen, N.~Wang, and Z.~Zhang.
\newblock Scale-aware trident networks for object detection.
\newblock In {\em IEEE/CVF International Conference on Computer Vision (ICCV)},
  pp. 6053--6062, 2019.

\bibitem{lin2017focal}
T.~{Lin}, P.~{Goyal}, R.~{Girshick}, K.~{He}, and P.~{Dollár}.
\newblock Focal loss for dense object detection.
\newblock {\em IEEE Transactions on Pattern Analysis and Machine Intelligence},
  42(2):318--327, 2020.

\bibitem{lin2017feature}
T.-Y. Lin, P.~Doll{\'a}r, R.~Girshick, K.~He, B.~Hariharan, and S.~Belongie.
\newblock Feature pyramid networks for object detection.
\newblock In {\em IEEE Conference on Computer Vision and Pattern Recognition
  (CVPR)}, pp. 936--944, 2017.

\bibitem{lin2014microsoft}
T.-Y. Lin, M.~Maire, S.~Belongie, J.~Hays, P.~Perona, D.~Ramanan,
  P.~Doll{\'a}r, and C.~L. Zitnick.
\newblock Microsoft {COCO}: Common objects in context.
\newblock In {\em European Conference on Computer Vision (ECCV)}, pp. 740--755.
  Springer, 2014.

\bibitem{liu2018mobile}
M.~Liu and M.~Zhu.
\newblock Mobile video object detection with temporally-aware feature maps.
\newblock In {\em IEEE/CVF Conference on Computer Vision and Pattern
  Recognition (CVPR)}, pp. 5686--5695, 2018.

\bibitem{liu2019towards}
R.~Liu, J.~Zhang, S.~Chen, and C.~Arth.
\newblock Towards {SLAM}-based outdoor localization using poor {GPS} and 2.5{D}
  building models.
\newblock In {\em 2019 IEEE International Symposium on Mixed and Augmented
  Reality (ISMAR)}, pp. 1--7, 2019.

\bibitem{liu2016ssd}
W.~Liu, D.~Anguelov, D.~Erhan, C.~Szegedy, S.~Reed, C.-Y. Fu, and A.~C. Berg.
\newblock {SSD}: Single shot multibox detector.
\newblock In {\em European Conference on Computer Vision (ECCV)}, pp. 21--37.
  Springer, 2016.

\bibitem{mccormac2017semanticfusion}
J.~McCormac, A.~Handa, A.~Davison, and S.~Leutenegger.
\newblock Semantic{F}usion: Dense 3{D} semantic mapping with convolutional
  neural networks.
\newblock In {\em International Conference on Robotics and Automation (ICRA)},
  pp. 4628--4635, 2017.

\bibitem{mu2016slam}
B.~Mu, S.-Y. Liu, L.~Paull, J.~Leonard, and J.~P. How.
\newblock {SLAM} with objects using a nonparametric pose graph.
\newblock In {\em IEEE/RSJ International Conference on Intelligent Robots and
  Systems (IROS)}, pp. 4602--4609, 2016.

\bibitem{indoorboundary}
A.~Phalak, Z.~Chen, D.~Yi, K.~Gupta, V.~Badrinarayanan, and A.~Rabinovich.
\newblock Deep{P}erimeter: Indoor boundary estimation from posed monocular
  sequences.
\newblock {\em arXiv preprint arXiv:1904.11595}, 2019.

\bibitem{redmon2016you}
J.~Redmon, S.~Divvala, R.~Girshick, and A.~Farhadi.
\newblock You only look once: unified, real-time object detection.
\newblock In {\em IEEE Conference on Computer Vision and Pattern Recognition
  (CVPR)}, pp. 779--788, 2016.

\bibitem{redmon2017yolo9000}
J.~Redmon and A.~Farhadi.
\newblock {YOLO9000}: better, faster, stronger.
\newblock In {\em IEEE Conference on Computer Vision and Pattern Recognition
  (CVPR)}, pp. 6517--6525, 2017.

\bibitem{ren2015faster}
S.~{Ren}, K.~{He}, R.~{Girshick}, and J.~{Sun}.
\newblock Faster {R-CNN}: Towards real-time object detection with region
  proposal networks.
\newblock {\em IEEE Transactions on Pattern Analysis and Machine Intelligence},
  39(6):1137--1149, 2017.

\bibitem{runz2018maskfusion}
M.~Runz, M.~Buffier, and L.~Agapito.
\newblock Mask{F}usion: Real-time recognition, tracking and reconstruction of
  multiple moving objects.
\newblock In {\em IEEE International Symposium on Mixed and Augmented Reality
  (ISMAR)}, pp. 10--20, 2018.

\bibitem{salas2013slam++}
R.~F. Salas-Moreno, R.~A. Newcombe, H.~Strasdat, P.~H. Kelly, and A.~J.
  Davison.
\newblock {SLAM}++: Simultaneous localisation and mapping at the level of
  objects.
\newblock In {\em IEEE Conference on Computer Vision and Pattern Recognition
  (CVPR)}, pp. 1352--1359, 2013.

\bibitem{singh2018sniper}
B.~Singh, M.~Najibi, and L.~S. Davis.
\newblock {SNIPER}: Efficient multi-scale training.
\newblock In {\em 32nd Conference on Neural Information Processing Systems
  (NeurIPS)}, pp. 9310--9320, 2018.

\bibitem{sunderhauf2017meaningful}
N.~S{\"u}nderhauf, T.~T. Pham, Y.~Latif, M.~Milford, and I.~Reid.
\newblock Meaningful maps with object-oriented semantic mapping.
\newblock In {\em IEEE/RSJ International Conference on Intelligent Robots and
  Systems (IROS)}, pp. 5079--5085, 2017.

\bibitem{tekin2018real}
B.~Tekin, S.~N. Sinha, and P.~Fua.
\newblock Real-time seamless single shot 6{D} object pose prediction.
\newblock In {\em IEEE/CVF Conference on Computer Vision and Pattern
  Recognition (CVPR)}, pp. 292--301, 2018.

\bibitem{tian2019occlusion}
Y.~Tian, Y.~Ma, S.~Quan, and Y.~Xu.
\newblock Occlusion and collision aware smartphone {AR} using time-of-flight
  camera.
\newblock In {\em International Symposium on Visual Computing (ISVC)}, pp.
  141--153. Springer, 2019.

\bibitem{tremblay2018deep}
J.~Tremblay, T.~To, B.~Sundaralingam, Y.~Xiang, D.~Fox, and S.~Birchfield.
\newblock Deep object pose estimation for semantic robotic grasping of
  household objects.
\newblock In {\em Conference on Robot Learning (CoRL)}, 2018.

\bibitem{uchiyama2012object}
H.~Uchiyama and E.~Marchand.
\newblock Object detection and pose tracking for augmented reality: recent
  approaches.
\newblock In {\em 18th Korea-Japan Joint Workshop on Frontiers of Computer
  Vision}, 2012.

\bibitem{valentin2018depth}
J.~Valentin, A.~Kowdle, J.~T. Barron, N.~Wadhwa, M.~Dzitsiuk, M.~Schoenberg,
  V.~Verma, A.~Csaszar, E.~Turner, I.~Dryanovski, J.~Afonso, J.~Pascoal,
  K.~Tsotsos, M.~Leung, M.~Schmidt, O.~Guleryuz, S.~Khamis, V.~Tankovitch,
  S.~Fanello, S.~Izadi, and C.~Rhemann.
\newblock Depth from motion for smartphone {AR}.
\newblock {\em ACM Transactions on Graphics (TOG)}, 37(6), Dec. 2018.

\bibitem{wang2019densefusion}
C.~Wang, D.~Xu, Y.~Zhu, R.~Mart{\'\i}n-Mart{\'\i}n, C.~Lu, L.~Fei-Fei, and
  S.~Savarese.
\newblock Dense{F}usion: 6{D} object pose estimation by iterative dense fusion.
\newblock In {\em IEEE/CVF Conference on Computer Vision and Pattern
  Recognition (CVPR)}, pp. 3338--3347, 2019.

\bibitem{wong2019yolo}
A.~Wong, M.~Famuori, M.~J. Shafiee, F.~Li, B.~Chwyl, and J.~Chung.
\newblock {YOLO} {N}ano: a highly compact you only look once convolutional
  neural network for object detection.
\newblock {\em arXiv preprint arXiv:1910.01271}, 2019.

\bibitem{xiang2015data}
Y.~Xiang, W.~Choi, Y.~Lin, and S.~Savarese.
\newblock Data-driven 3{D} voxel patterns for object category recognition.
\newblock In {\em IEEE Conference on Computer Vision and Pattern Recognition
  (CVPR)}, pp. 1903--1911, 2015.

\bibitem{xu2018multi}
Y.~Xu, Y.~Wu, and H.~Zhou.
\newblock Multi-scale voxel hashing and efficient 3{D} representation for
  mobile augmented reality.
\newblock In {\em IEEE/CVF Conference on Computer Vision and Pattern
  Recognition Workshops (CVPRW)}, pp. 1618--1625, 2018.

\bibitem{yu2019variational}
H.~Yu, J.~Moon, and B.~Lee.
\newblock A variational observation model of 3{D} object for probabilistic
  semantic {SLAM}.
\newblock In {\em International Conference on Robotics and Automation (ICRA)},
  pp. 5866--5872, 2019.

\bibitem{zhang2019hierarchical}
J.~Zhang, M.~Gui, Q.~Wang, R.~Liu, J.~Xu, and S.~Chen.
\newblock Hierarchical topic model based object association for semantic
  {SLAM}.
\newblock {\em IEEE Transactions on Visualization and Computer Graphics
  (TVCG)}, 25(11):3052--3062, 2019.

\bibitem{zhu2015segdeepm}
Y.~Zhu, R.~Urtasun, R.~Salakhutdinov, and S.~Fidler.
\newblock seg{D}eep{M}: Exploiting segmentation and context in deep neural
  networks for object detection.
\newblock In {\em IEEE Conference on Computer Vision and Pattern Recognition
  (CVPR)}, pp. 4703--4711, 2015.

\end{thebibliography}
\end{document}